\documentclass{article}
\usepackage{iclr2027_conference,times}
\usepackage{amsmath,amssymb,booktabs,array,tabularx,multirow}
\usepackage{graphicx}
\usepackage[table]{xcolor}
\usepackage{caption}
\usepackage{needspace}
\usepackage[hyperfootnotes=false]{hyperref}
\usepackage{url}
\newcolumntype{C}{>{\centering\arraybackslash}X}
\newcolumntype{L}{>{\raggedright\arraybackslash}X}
\definecolor{tablehead}{HTML}{ECECEC}
\definecolor{tablepanel}{HTML}{F6F6F6}
\definecolor{tablemuted}{HTML}{777777}

\newcommand{\addon}[1]{\hspace*{0.75em}+ #1}

\newcommand{\methodname}{FocusOPD}
\newcommand{\papertitle}{FocusOPD: Structure-Guided On-Policy \texorpdfstring{\\}{ } Distillation  for Diffusion Models}

\hypersetup{hidelinks,pdftitle={FocusOPD: Structure-Guided On-Policy Distillation for Diffusion Models}}

\title{\papertitle}
\author{
\begin{minipage}[t]{\dimexpr\textwidth-2\tabcolsep-1em\relax}
\raggedright
Xinrui Lin\textsuperscript{1,*}, Shumin Wang\textsuperscript{1,*}, Suqi Song\textsuperscript{2}, Fan Li\textsuperscript{3,\ensuremath{\dagger}}, Yanyong Zhang\textsuperscript{1,\ensuremath{\dagger}}\\[3pt]
\normalfont\small
\textsuperscript{1}University of Science and Technology of China, Hefei, China\\
\textsuperscript{2}The Hong Kong University of Science and Technology (Guangzhou)\\
\textsuperscript{3}Huawei Noah's Ark Lab\\
\texttt{\{xinruilin,shuminwang\}@mail.ustc.edu.cn}\\
\texttt{ssong943@connect.hkust-gz.edu.cn}\\
\texttt{lifan61@huawei.com}, \texttt{yanyongz@ustc.edu.cn}
\end{minipage}
}
\iclrfinalcopy
\date{}
\hypersetup{pdfauthor={Xinrui Lin, Shumin Wang, Suqi Song, Fan Li, Yanyong Zhang}}
\begin{document}
\maketitle
\fancyhead{}
\renewcommand{\headrulewidth}{0pt}
\begingroup
\renewcommand{\thefootnote}{*}
\footnotetext{Equal contribution.}
\renewcommand{\thefootnote}{\ensuremath{\dagger}}
\footnotetext{Corresponding authors.}
\endgroup

\begin{abstract}
Multi-teacher on-policy distillation enables rectified-flow models with
diffusion-transformer architectures to combine task-specific capabilities
within a single student. However, matching model outputs does not guarantee
that their internal representations are aligned, and it remains unclear where
and when additional feature supervision is most useful. 
We analyze the teacher–student feature mismatch across Transformer blocks and denoising timesteps, and find that the mismatch is concentrated in deeper Transformer blocks during early denoising and persists after OPD distillation, indicating ineffective distillation. 
Motivated
by this observation, we introduce \methodname{}, which augments the base OPD objective with token-wise directional feature alignment restricted to deep layers and early denoising steps. 
Across two
rectified-flow backbones and multiple OPD objectives, \methodname{} improves capability transfer in compositional generation, text rendering, and
preference-based evaluation. 
On the
primary benchmark, it improves over DiffusionOPD from 0.953 to 0.967 on
GenEval, from 23.797 to 23.944 on PickScore, and from 6.052 to 6.199 on
LAION-Aesthetic. Ablations show that selective deep and early supervision is more
effective and efficient than alignment over the full network and trajectory. 
\end{abstract}

\section{Introduction}
\label{sec:introduction}

On-policy distillation (OPD) transfers teacher behavior through feedback
on student-generated trajectories~\citep{agarwal2024onpolicy,lu2025onpolicydistillation}.
For diffusion models, multi-teacher OPD integrates specialized capabilities,
including compositional generation, text rendering, and alignment with
human preferences, into a single student~\citep{ma2026mopd,diffusionopd,flowopd}.
Recent methods select denoising states for teacher feedback and align
representation changes across network blocks~\citep{danceopd,stepopd}.
Yet these developments leave open how the value of supervision varies
across blocks and denoising stages. We therefore ask:
\emph{what patterns characterize the utility of additional feature
supervision across network depth and denoising steps?}

To investigate this question, we begin by examining teacher--student
feature mismatch in SD3.5-M~\cite{sd3} and Z-Image~\citep{zimage}.
Evaluating teacher and student on the same noisy inputs, we find that
feature mismatch is generally larger in deeper blocks and during early,
high-noise denoising. Its relative organization remains qualitatively
similar before and after output-level OPD, even though its magnitude
may change. We call this a \emph{persistent depth--temporal structure}.
The marginal trends locate larger discrepancies; the joint profiles
reveal that their organization persists through training.
Neither observation alone explains the underlying mechanism.

We examine two mechanisms consistent with this structure.
Two lemmas characterize how local discrepancies propagate across network
blocks and how weaker observations preserve prior-dependent differences
in a Gaussian denoising model.
Building on this account, we derive a local criterion for selective
feature supervision: under local smoothness and gradient-compatibility
conditions, allocating a fixed auxiliary weight to sufficiently larger
discrepancies can yield greater output-loss reduction for small auxiliary
updates.
Two lightweight diagnostics support the proposed mechanisms:
upstream differences dominate mismatch at selected deep blocks, and
noise-dependent scaling improves held-out prediction over a timestep
control.

Motivated by this analysis, we introduce \methodname{}, a
straightforward way to use the persistent depth--temporal structure in
multi-teacher OPD (MOPD). It augments the base output objective with tokenwise
cosine alignment of conditional image features in deep layers during
early denoising. Output and feature queries are specified independently,
and inference requires no additional components. On SD3.5-M with
DiffusionOPD, it improves GenEval from 0.953 to 0.967, OCR accuracy from
0.958 to 0.960, and PickScore from 23.797 to 23.944, and ImageReward from 1.471 to
1.517. The ablations show that gains depend on supervision placement: Last-12/Front-40\% reaches LAION-Aesthetic 6.214 with a 432 MiB feature cache,
whereas full-trajectory supervision yieds a GenEval score of 0.951, and supervising all blocks raises it only to 0.965 while doubling the cache.
Our main contributions are:
\begin{itemize}
    \item \textbf{Observation}
    (Section~\ref{sec:mismatch_observation}).
    We identify persistent structure of teacher-student mismatch across depth and denoising
    time in the MOPD of SD3.5-M and Z-Image.

    \item \textbf{Theory \& Evidence}
(Sections~\ref{sec:mechanism}--\ref{sec:validation}).
We characterize two mismatch mechanisms, support them with targeted
diagnostics, and derive a conditional local criterion for selective
supervision.

    \item \textbf{Method}
    (Section~\ref{sec:method}).
    We introduce \methodname{}, aligning deep features in early denoising steps while retaining the base OPD protocol.

    \item \textbf{Experiments \& Insights}
    (Section~\ref{sec:experiments}).
        We improve capability transfer across composition, text rendering, and
    preference metrics. The ablations show that selective supervision of deep
    layers during early denoising is effective, while broader temporal or
    depth coverage adds cost without consistent gains.
\end{itemize}

\label{sec:mismatch_observation}
\begin{figure}[t]
    \centering
    \includegraphics{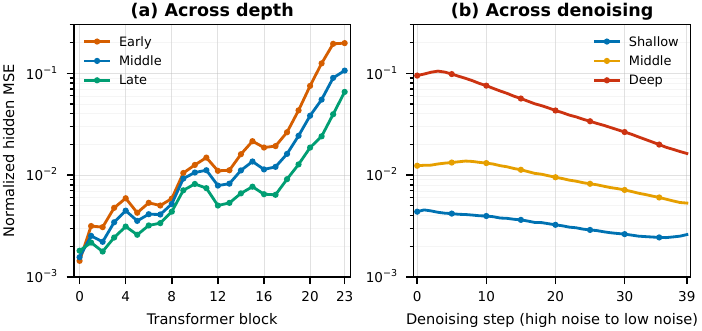}\vspace{-5pt}
    \caption{Teacher--student feature MSE across (a) transformer blocks
        and (b) denoising steps, ordered from high to low noise. Different curves indicates different groups of (a)denoising steps (b)blocks.}
    \label{fig:mismatch_curves}
\end{figure}
\section{A Persistent Structure of Teacher--Student Mismatch}

Output-level OPD transfers teacher behavior by matching denoising predictions. To understand the internal discrepancies that remain under this supervision, we begin by examining where teacher--student
representations differ across network depth and denoising time, and
whether their mismatch structure changes with training or varies on different models.

On SD3.5-M, we measure the MSE between corresponding
conditional image-token features, evaluating teacher and student on the same noisy latent, timestep, and prompt.
Figure~\ref{fig:mismatch_curves} shows that mismatch is larger in deeper transformer blocks and during early, high-noise denoising steps.

These trends identify where discrepancies are larger, but do not explain how they arise. To further characterize this pattern, we examine its
joint distribution across blocks and denoising steps before and after output-level OPD and extend the experiments with Z-Image.

\begin{figure}[t]
    \centering
    \includegraphics[width=\linewidth,height=1.10in,keepaspectratio]{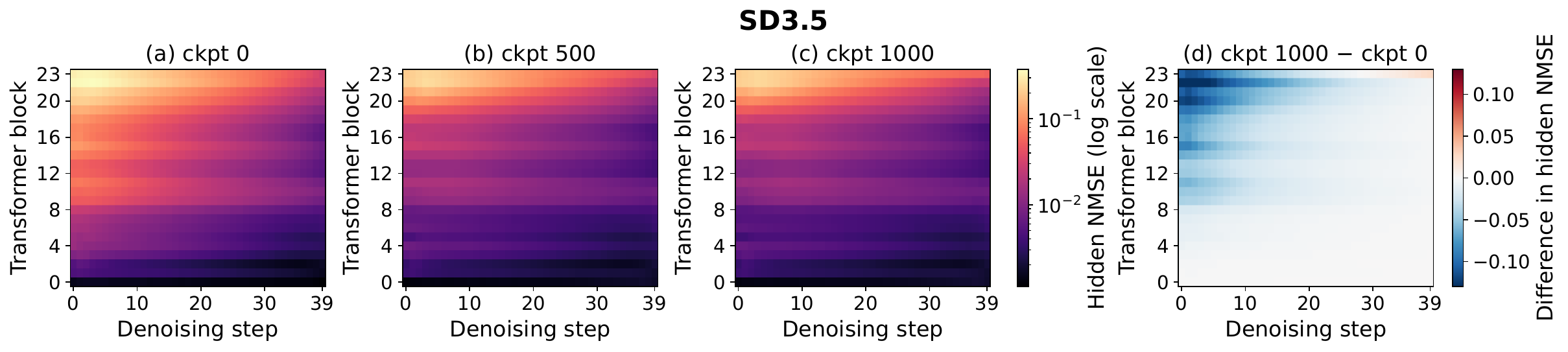}

    \vspace{1pt}
    \includegraphics[width=\linewidth,height=1.10in,keepaspectratio]{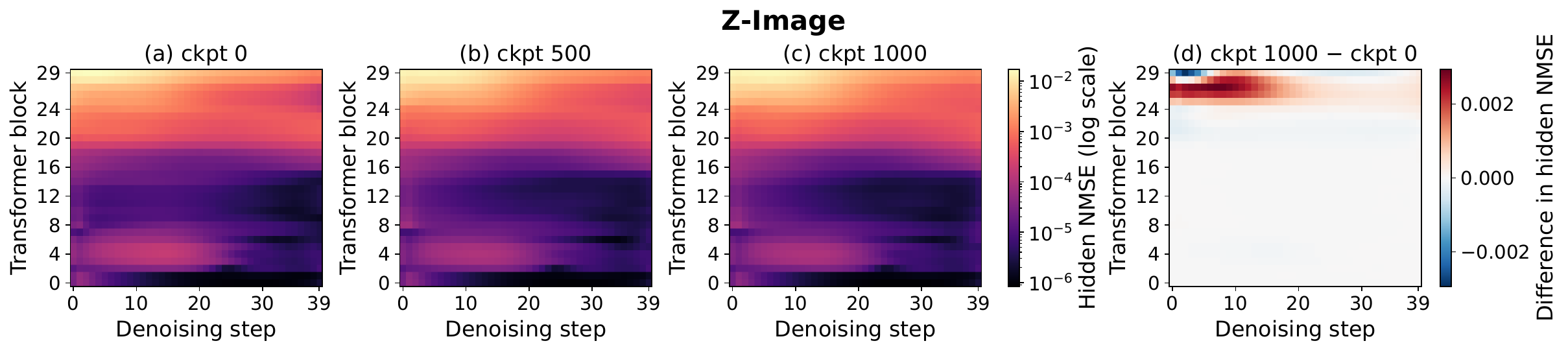}

    \caption{Hidden-NMSE heatmaps. Columns show checkpoints 0, 500, 1000, and
        the checkpoint-1000 minus checkpoint-0 difference.}
    \label{fig:mismatch_heatmaps}
\end{figure}

Figure~\ref{fig:mismatch_heatmaps} reveals a consistent pattern across
both backbones: larger discrepancies are concentrated in deeper blocks
and earlier denoising steps. This organization persists after training,
with only small differences between training steps 0 and 1000, as shown
in the rightmost column.
We refer to this recurring pattern as a \emph{persistent depth--temporal
structure}. Persistence describes the relative distribution of mismatch,
rather than an unchanged absolute MSE. This raises the question:
\emph{What gives rise to this structure, and why does it persist after
output-level distillation?} Section~\ref{sec:mechanism} develops a
mechanistic account of these observations.

\section{From Mismatch Mechanisms to Selective Supervision}
\label{sec:mechanism}

We first examine two mechanisms consistent with the observed structure,
then analyze when mismatch can guide feature supervision.
All proofs are in Appendix~\ref{app:mismatch_theory}.

\textbf{Notation.}\;
Fix a student $S$ and teacher $T$ at the same noisy latent $x_s$ and
text condition $c$, with $s$ indexing denoising from high to low noise.
For $M\in\{S,T\}$, let $z_{\ell,s}^M$ be the complete input to block
$\ell$, including image/text tokens and side inputs, and $F_{\ell,s}^M$
its block map. Thus $z_{\ell+1,s}^M=F_{\ell,s}^M(z_{\ell,s}^M)$;
write $d_{\ell,s}=z_{\ell,s}^S-z_{\ell,s}^T$.

\begin{lemma}[Depth-wise discrepancy transport]
\label{lem:discrepancy_propagation}
For continuously differentiable student blocks, define
$b_{j,s}=F_{j,s}^S(z_{j,s}^T)-F_{j,s}^T(z_{j,s}^T)$.
Let $A_{j,s}$ be the student Jacobian averaged along the segment
joining its teacher and student inputs, and set
$P_{\ell\leftarrow j,s}=A_{\ell-1,s}\cdots A_{j,s}$,
$P_{\ell\leftarrow\ell,s}=I$. Then
\begin{equation}
    d_{\ell,s}=P_{\ell\leftarrow0,s}d_{0,s}
    +\sum_{j=0}^{\ell-1}P_{\ell\leftarrow j+1,s}b_{j,s}.
    \label{eq:discrepancy_expansion}
\end{equation}
\end{lemma}
Deep mismatch thus combines local discrepancies with differences
inherited from upstream states; their interaction determines accumulation.

\begin{lemma}[Noise-dependent prior sensitivity]
\label{lem:prior_sensitivity}
Let $y\mid c,M\sim\mathcal N(\mu_M(c),\tau^2I)$ be a clean-latent
prior with mean $\mu_M(c)$ and shared variance $\tau^2>0$.
Observe $x_s=a_sy+\sigma_s\varepsilon$, with signal coefficient $a_s$,
noise scale $\sigma_s>0$, and $\varepsilon\sim\mathcal N(0,I)$
independent of $y$. The posterior means
$\widehat y_M=\mathbb E_M[y\mid x_s,c]$ satisfy
\begin{equation}
    \widehat y_S-\widehat y_T
    =\rho_s\bigl(\mu_S(c)-\mu_T(c)\bigr),
    \qquad
    \rho_s=\frac{\sigma_s^2}{a_s^2\tau^2+\sigma_s^2}.
    \label{eq:prior_retention}
\end{equation}
\end{lemma}
The priors encode model-specific expectations under the same text.
Lower $\mathrm{SNR}_s=a_s^2/\sigma_s^2$ preserves more of this prior
difference. Together, Lemmas~\ref{lem:discrepancy_propagation}
and~\ref{lem:prior_sensitivity} motivate a structured view of mismatch
across depth and denoising time. Building on this account,
Proposition~\ref{prop:selective_update} establishes a conditional implication
for feature supervision: under local smoothness and gradient-compatibility
conditions, sufficiently larger mismatch can yield a greater single-step
reduction in output loss at a matched supervision budget.

\paragraph{Local supervision utility.}
Fix a replay batch and teacher targets for one update comparison.
For position $i=(\ell,s)$, let $\phi_i^M$ stack the image features,
with averaging factors and optional token normalization included.
For student parameters $\theta$, define
$e_i(\theta)=\phi_i^S(\theta)-\phi_i^T$,
$D_i=\|e_i\|_2^2$, and $\ell_i=D_i/2$.
Let $g=\nabla_\theta\mathcal L_{\mathrm{out}}$ be the output-loss
gradient and $g_i=\nabla_\theta\ell_i$ the feature gradient, all evaluated
at $\theta$. A supervision distribution $p_i\geq0$, $\sum_i p_i=1$,
defines $g_p=\sum_i p_i g_i$ and mean mismatch $S_p=\sum_i p_iD_i$.
For a fixed learning rate $\eta>0$ and auxiliary weight $\lambda>0$,
compare
\begin{equation}
    \theta_{\mathrm{out}}^+=\theta-\eta g,
    \qquad
    \theta_p^+=\theta_{\mathrm{out}}^+-\delta g_p,
    \qquad \delta=\eta\lambda.
    \label{eq:local_joint_update}
\end{equation}
Write $g_{\mathrm{out}}^+=\nabla\mathcal L_{\mathrm{out}}
(\theta_{\mathrm{out}}^+)$ and define the correction efficiency
$\alpha_i=\langle g_{\mathrm{out}}^+,g_i\rangle/D_i$ for $D_i>0$,
with $\alpha_i=0$ otherwise. It measures output correction per unit
feature mismatch.

\begin{proposition}[Local utility of selective supervision]
\label{prop:selective_update}
Suppose the features are differentiable and the output-loss gradient
is $L_{\mathrm{sm}}$-Lipschitz on the compared update segments.
The additional output-loss reduction
$\mathcal G(p)=\mathcal L_{\mathrm{out}}(\theta_{\mathrm{out}}^+)
-\mathcal L_{\mathrm{out}}(\theta_p^+)$ satisfies
\begin{equation}
    \mathcal G(p)=\delta\sum_i p_i\alpha_iD_i+R_p,
    \qquad
    |R_p|\leq\frac{L_{\mathrm{sm}}\delta^2}{2}\|g_p\|_2^2.
    \label{eq:selective_output_gain}
\end{equation}
For two distributions $p,q$, assume
$0<\alpha_{\min}\leq\alpha_i\leq\alpha_{\max}$ at their
positive-mismatch positions. If
$\alpha_{\min}S_p>\alpha_{\max}S_q$, then
$\mathcal G(p)>\mathcal G(q)$ for sufficiently small common $\delta>0$.
\end{proposition}

The proposition links supervision utility to mismatch and its
compatibility with the output objective. When correction efficiencies
are positive and comparable, it favors regions whose larger mismatch
outweighs the efficiency variation. Combined with the observed
concentration in deep layers and early steps, this motivates the
selective design in Section~\ref{sec:method}.
For cosine supervision, $D_i$ is the corresponding unit-feature
discrepancy (Appendix~\ref{app:angular_response}).

\begin{figure}[t]
    \centering
    \includegraphics[width=\linewidth]{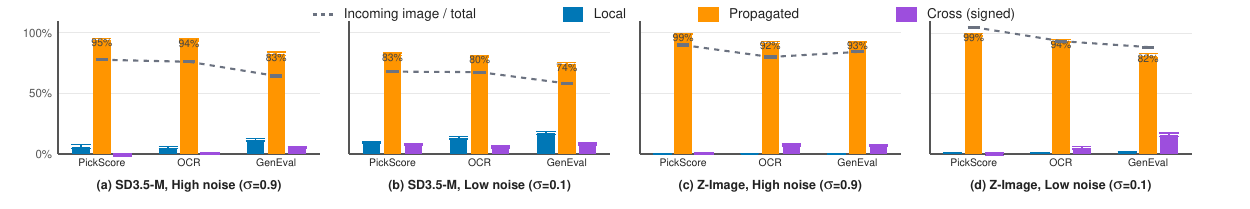}
    \caption{Deep-block decomposition across three teachers. Terms are
        normalized by total mean discrepancy energy. Block indices are
        18 for SD3.5-M and 22 for Z-Image.
        Error bars: 95\% prompt-bootstrap intervals.}
    \label{fig:depth-decomposition}
\end{figure}

\section{Empirical Evidence for the Mismatch Mechanisms}
\label{sec:validation}
\begin{wraptable}[10]{r}{0.48\textwidth}
    \centering
    \caption{Held-out prediction NMSE. Lower is better.}
    \label{tab:noise-scaling}
    \small
    \setlength{\tabcolsep}{3pt}
    \begin{tabular}{@{}llrr@{}}
        \toprule
        Model & Teacher & $t$-ctrl. & Scaling \\
        \midrule
        SD3.5-M & PickScore & 1.303 & \textbf{0.748} \\
                & OCR       & 0.728 & \textbf{0.547} \\
                & GenEval   & 5.932 & \textbf{0.879} \\
        \midrule
        Z-Image & PickScore & 1.938 & \textbf{0.947} \\
                & OCR       & 3.380 & \textbf{0.786} \\
                & GenEval   & 8.243 & \textbf{0.862} \\
        \bottomrule
    \end{tabular}
\end{wraptable}
We examine Lemmas~\ref{lem:discrepancy_propagation}
and~\ref{lem:prior_sensitivity} using output-only DiffusionOPD
checkpoint-1000 on SD3.5-M and Z-Image, each paired with PickScore,
OCR, and GenEval teachers. We use shared teacher--student inputs for
32 prompts per pair, fixing each prompt's clean latent and noise across $\sigma\in\{0.1,0.3,0.5,0.7,0.9\}$.

Evaluating student blocks on their native and complete teacher inputs
separates the local discrepancy $b_{\ell,s}$ from the propagated term
$p_{\ell,s}=d_{\ell+1,s}-b_{\ell,s}$.
We measure image-token energies and their signed interaction
(Equation~\eqref{eq:discrepancy_cross_term}). At the selected deep
positions, propagated energy is dominates the discrepancy
energy (Figure~\ref{fig:depth-decomposition}). Upstream state differences thus dominate mismatch at these positions. More results are in Appendix~\ref{app:evidence}.

To examine noise dependence, let
$\Delta_s=\widehat y_S-\widehat y_T$ denote the clean-latent estimate
difference, with $\star$ marking the reference at $\sigma=0.9$.
We compare the prediction $(\rho_s/\rho_\star)\Delta_\star$ from
Lemma~\ref{lem:prior_sensitivity} with the $t$-control
$(t_s/t_\star)\Delta_\star$. Here $t_s$ is the coefficient in
$\widehat y_M=x_s-t_s\widehat v_M$, where $\widehat v_M$ is the predicted
flow velocity. For each pair, we fit the single parameter $\tau^2$ on
16 prompts and evaluate normalized mean squared prediction error
(NMSE) on the remaining 16. Noise-dependent scaling yields lower
held-out NMSE in all six pairs (Table~\ref{tab:noise-scaling}). This result supports its predictive value beyond the output-conversion factor alone.

\section{\methodname: Structure-Guided Feature Distillation}
\label{sec:method}

\begin{figure}[t]
    \centering
    \includegraphics[width=0.99\linewidth]{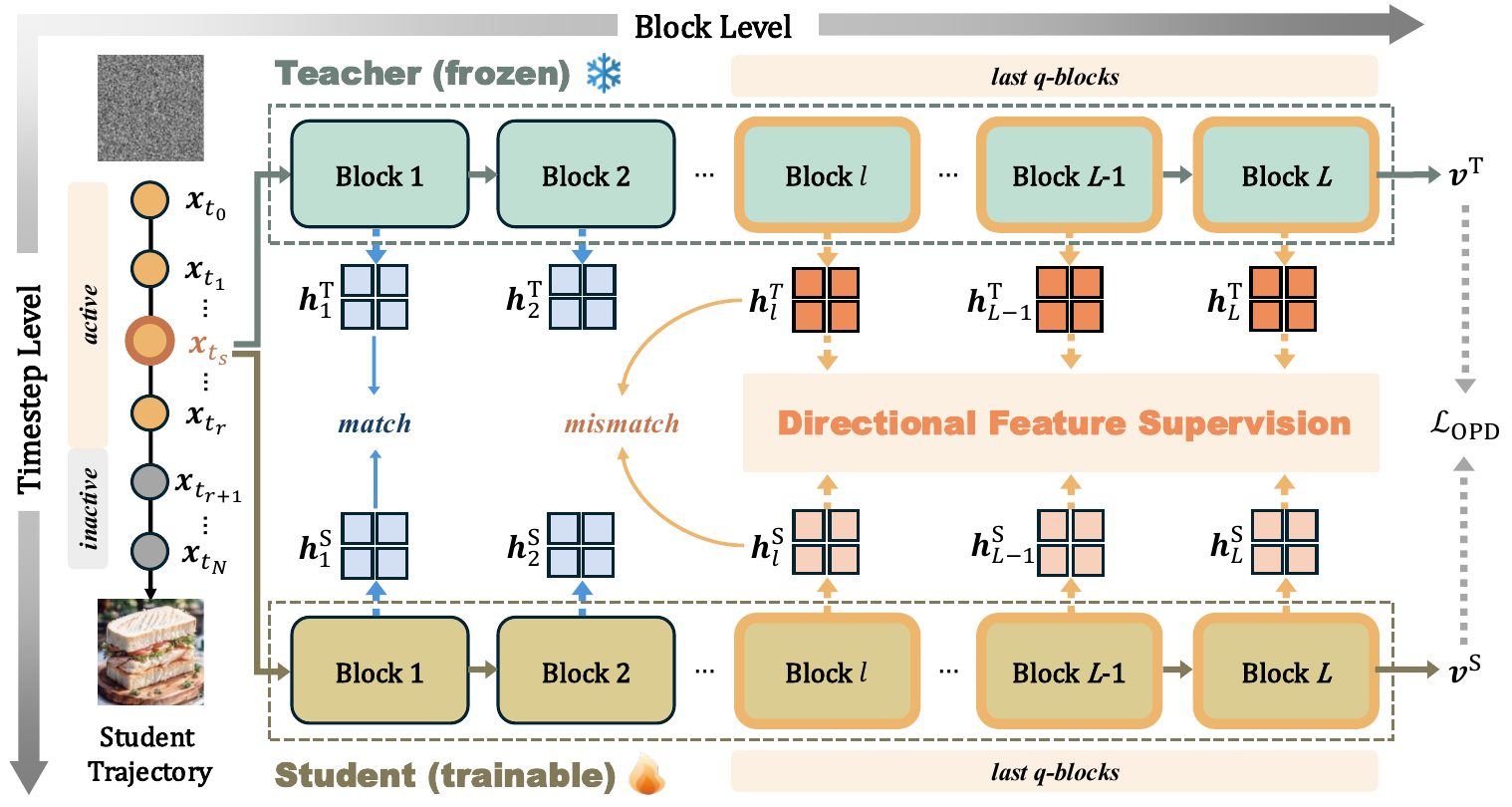}
    \caption{Overview of \methodname{}.
        Teacher and student are evaluated on shared states from student
        trajectories. Output supervision follows the base OPD protocol,
        while directional feature supervision targets deep layers during
        early denoising. Only the trained student is used at inference.}
    \label{fig:method}
\end{figure}

Motivated by Proposition~\ref{prop:selective_update}, we introduce
\methodname{}, a straightforward way to use the persistent
depth--temporal mismatch structure in multi-teacher OPD.
We concentrate auxiliary supervision on deep layers and early denoising
steps, where the observed mismatch is concentrated. The method augments
output-level distillation with directional feature alignment in this
fixed region (Figure~\ref{fig:method}).

\subsection{Distillation on Student Trajectories}
\label{sec:method-output}

Let $S_\theta$ be the student and $\{T_m\}_{m=1}^{K}$ fixed
task-specific teachers sharing its backbone architecture.
We sample task $m$ from a distribution $\pi$ and prompt $c$ from its
prompt set $\mathcal C_m$. At each round, the student with parameters
$\theta_{\mathrm{roll}}$ generates an $N$-step trajectory
$\tau=(x_{t_0},\ldots,x_{t_N})$, where $t_0>\cdots>t_N$.
Optimization uses detached states
$\bar x_{t_s}=\operatorname{sg}[x_{t_s}]$ for $s=0,\ldots,N-1$,
with $\operatorname{sg}$ denoting stop-gradient.
Teacher $T_m$ and student $S_\theta$ receive the same
$(\bar x_{t_s},t_s,c)$.
We retain the base output objective $\mathcal L_{\mathrm{OPD}}$,
including its targets, weights, and query set
$\mathcal I_{\mathrm{out}}\subseteq\{0,\ldots,N-1\}$.

\subsection{Selecting the Supervision Region}
\label{sec:method-selection}

We select a contiguous deep-layer window and an early denoising
window. For a backbone with $L$ blocks, let $q\in\{1,\ldots,L\}$
be the number of selected blocks. Their output positions are
\begin{equation}
    \mathcal D_q=\{L-q+1,\ldots,L\}.
    \label{eq:depth-window}
\end{equation}
For an early-window fraction $r\in[1/N,1]$, define
\begin{equation}
    m_r=\lfloor Nr\rfloor,
    \qquad
    \mathcal I_{\mathrm{feat}}(r)=\{0,\ldots,m_r-1\}.
    \label{eq:time-window}
\end{equation}
The resulting region is
$\mathcal D_q\times\mathcal I_{\mathrm{feat}}(r)$, with early steps
measured from the high-noise start of each trajectory.
Its use as a fixed region is motivated by the observed persistence
and Proposition~\ref{prop:selective_update}.
Feature queries are specified independently of
$\mathcal I_{\mathrm{out}}$, so output supervision continues according
to the base protocol beyond the feature window.

\subsection{Directional Feature Supervision}
\label{sec:method-features}
\label{sec:method-training}

For $M\in\{S_\theta,T_m\}$, let
$h_{\ell,s,p}^{M}\in\mathbb R^{d_h}$ be image token $p$ after
$\ell$ blocks at $(\bar x_{t_s},t_s,c)$, where $d_h$ is the feature
dimension and $P$ is the number of image tokens.
We extract features from the text-conditional branch; output predictions
use the base guidance configuration.
With $\widehat h=h/\max(\|h\|_2,\epsilon)$ and norm floor
$\epsilon>0$, the feature loss at step $s$ is
\begin{equation}
    \mathcal L_{\mathrm{feat}}^{m,s}
    =
    \frac{1}{|\mathcal D_q|P}
    \sum_{\mathcal L\in\mathcal D_q}\sum_{p=1}^{P}
    \left(
        1-
        \left\langle
            \widehat h_{\ell,s,p}^{S_\theta},
            \operatorname{sg}\!\left[
                \widehat h_{\ell,s,p}^{T_m}
            \right]
        \right\rangle
    \right).
    \label{eq:feature-loss}
\end{equation}
This objective aligns feature directions, averaging over selected
blocks and image tokens. Its normalized-feature formulation allows
Proposition~\ref{prop:selective_update} to apply directly to
cosine-loss gradients (ref Appendix~\ref{app:angular_response} for details).

The joint training objective is
\begin{equation}
    \mathcal L(\theta)
    =
    \mathcal L_{\mathrm{OPD}}(\theta)
    +
    \lambda\,
    \mathbb E_{m,c,\tau}
    \left[
        \frac{1}{N}
        \sum_{s\in\mathcal I_{\mathrm{feat}}(r)}
        \mathcal L_{\mathrm{feat}}^{m,s}
    \right],
    \label{eq:joint-objective}
\end{equation}
where $\tau\sim S_{\theta_{\mathrm{roll}}}(\cdot\mid c)$ and
$\lambda\geq0$ controls feature supervision.
The auxiliary loss is normalized by the full trajectory length $N$.
We keep $(q,r,\lambda)$ fixed during training and shared across task
teachers; concrete settings and supervision-budget comparisons are
reported in Section~\ref{sec:experiments}.

Training replays states in
$\mathcal I_{\mathrm{out}}\cup\mathcal I_{\mathrm{feat}}(r)$,
applying each loss at its designated queries and sharing teacher
forward passes where queries overlap.
Only student parameters are updated, with fresh student trajectories
collected in subsequent rounds. Inference uses the trained student
without additional components.

\section{Experiments}
\label{sec:experiments}

\subsection{Experimental Setup}
\label{sec:exp_setup}
\textbf{Models and Evaluation.}
Following the multi-teacher setting of DiffusionOPD \citep{diffusionopd}, we evaluate two diffusion backbones, SD3.5-M \citep{sd3} and Z-Image \citep{zimage}, at a resolution of \(512\times512\). For each backbone, the student and three task-specific teachers are initialized from the same base model, with teachers specializing in compositional alignment, text rendering, and human preference optimization. For SD3.5-M, we use the teachers trained in DiffusionOPD, while for Z-Image we train three teachers using the same task-specific RL protocols. All students generate ten-step on-policy trajectories with a deterministic flow solver and only LoRA \citep{hu2022lora} parameters are optimized. All methods are trained for 1000 epochs.

We evaluate compositional alignment on GenEval \citep{geneval} and report its six category scores and overall score. Text rendering is evaluated on a held-out OCR split using OCR accuracy. Human preference and visual quality are evaluated on DrawBench \citep{imagen} with PickScore \citep{pickscore}, HPSv2.1 \citep{hps}, LAION-Aesthetic \citep{aesthetic}, and ImageReward \citep{imagereward}. We use the same evaluation prompts, resolution, sampler, guidance settings, random seeds, and evaluator versions for all methods within each backbone.

\textbf{Baselines.}
We compare the base model and single-task teachers with multi-teacher OPD methods. DiffusionOPD \citep{diffusionopd} and DanceOPD \citep{danceopd} represent output-level distillation, while STEP-OPD \citep{stepopd} additionally supervises internal representation dynamics. Our method builds on each OPD objective while preserving its output target and teacher-routing strategy.

\subsection{Main Results}
\label{sec:main_results}
Table~\ref{tab:main} compares each baseline with and without our feature
supervision: DiffusionOPD, DanceOPD, and STEP-OPD  are each
followed by a corresponding ``+ \methodname'' row. 
This paired presentation tests whether the proposed feature objective is useful across different OPD output objectives and query strategies. The SD3.5-M and Z-Image experiments use the same teacher roles and evaluation protocol, with depth and denoising windows calibrated per backbone.

\begin{table}[!ht]
\centering
\caption{Multi-teacher distillation on GenEval, OCR, and DrawBench.
The table reports all GenEval categories, OCR accuracy, and four independent
DrawBench quality metrics. All metrics are higher-is-better; Scores are reported at checkpoint 1000.
Gray rows highlight our method applied to each baseline.}
\label{tab:main}
\scriptsize
\setlength{\tabcolsep}{0.65pt}
\renewcommand{\arraystretch}{1.05}
\begin{tabularx}{\linewidth}{@{}l *{12}{C}@{}}
\toprule
\multirow{2}{*}[-2pt]{\textbf{Method}}
& \multicolumn{7}{c}{\textbf{GenEval}} & \multicolumn{1}{c}{\textbf{OCR}}
& \multicolumn{4}{c}{\textbf{DrawBench}}\\
\cmidrule(lr){2-8}\cmidrule(lr){9-9}\cmidrule(l){10-13}
& Sing. & Two & Cnt. & Col. & Pos. & C-Att. & Ovr.
& Acc. & Pick & Aes. & HPS & ImgR\\
\midrule
\multicolumn{13}{@{}l}{\textbf{SD3.5-M}}\\
\addlinespace[1pt]
\multicolumn{13}{@{}l}{\textit{Single-task teacher references}}\\
Base model & 0.963 & 0.889 & 0.613 & 0.809 & 0.250 & 0.570 & 0.682 & 0.675 & 22.411 & 5.388 & 0.281 & 0.901\\
GenEval teacher & 1.000 & 1.000 & 0.950 & 0.936 & 0.960 & 0.900 & 0.958 & 0.507 & 22.061 & 5.202 & 0.248 & 0.611\\
OCR teacher & 0.988 & 0.838 & 0.663 & 0.787 & 0.310 & 0.480 & 0.678 & 0.961 & 22.351 & 5.273 & 0.273 & 0.938\\
Preference teacher & 1.000 & 0.970 & 0.625 & 0.894 & 0.350 & 0.640 & 0.746 & 0.729 & 24.049 & 6.237 & 0.346 & 1.520\\
\multicolumn{13}{@{}l}{\textit{Multi-objective OPD methods}}\\
DanceOPD & 0.988 & \textbf{1.000} & 0.900 & \textbf{0.926} & 0.960 & 0.790 & 0.927 & 0.952 & 23.573 & 5.872 & 0.317 & 1.367\\
\addon{\textbf{\methodname\ (ours)}} & \textbf{1.000} & 0.990 & 0.900 & 0.904 & 0.960 & 0.830 & 0.931 & 0.945 & 23.736 & 5.986 & 0.326 & 1.441\\
STEP-OPD & \underline{1.000} & \underline{1.000} & \underline{0.975} & \underline{0.926} & 0.980 & \underline{0.900} & 0.963 & \underline{0.961} & \textbf{24.048} & \underline{6.255} & \textbf{0.345} & \underline{1.518}\\
\addon{\textbf{\methodname\ (ours)}} & \textbf{1.000} & \textbf{1.000} & \underline{0.975} & 0.915 & 0.980 & \textbf{0.930} & \underline{0.967} & \textbf{0.961} & \underline{24.025} & \textbf{6.260} & \underline{0.345} & \textbf{1.520}\\
DiffusionOPD & \textbf{1.000} & \textbf{1.000} & 0.925 & 0.915 & \textbf{0.990} & 0.890 & 0.953 & 0.958 & 23.797 & 6.052 & 0.333 & 1.471\\
\addon{\textbf{\methodname\ (ours)}} & \textbf{1.000} & \textbf{1.000} & \textbf{0.988} & \textbf{0.926} & \textbf{0.990} & \underline{0.900} & \textbf{0.967} & 0.960 & 23.944 & 6.199 & 0.340 & 1.517\\
\addlinespace[3pt]
\midrule
\addlinespace[1pt]
\multicolumn{13}{@{}l}{\textbf{Z-Image}}\\
\addlinespace[1pt]
\multicolumn{13}{@{}l}{\textit{Single-task teacher references}}\\
Base model & 0.988 & 0.889 & 0.625 & 0.777 & 0.380 & 0.480 & 0.690 & 0.779 & 22.177 & 5.233 & 0.270 & 0.780\\
GenEval teacher & 1.000 & 0.990 & 0.913 & 0.936 & 0.970 & 0.880 & 0.948 & 0.835 & 22.287 & 5.113 & 0.262 & 0.893\\
OCR teacher & 0.975 & 0.838 & 0.600 & 0.798 & 0.360 & 0.410 & 0.664 & 0.876 & 22.226 & 5.220 & 0.272 & 0.797\\
Preference teacher & 1.000 & 0.899 & 0.600 & 0.819 & 0.320 & 0.620 & 0.710 & 0.867 & 23.485 & 5.942 & 0.332 & 1.361\\
\multicolumn{13}{@{}l}{\textit{Multi-objective OPD methods}}\\
DanceOPD & 0.975 & 0.828 & 0.588 & 0.883 & 0.430 & 0.650 & 0.726 & 0.873 & 22.894 & 5.531 & 0.300 & 1.100\\
\addon{\textbf{\methodname\ (ours)}} & 0.975 & \underline{0.960} & 0.763 & 0.894 & \textbf{0.990} & 0.800 & 0.897 & 0.875 & 23.203 & 5.691 & 0.315 & 1.247\\
STEP-OPD & \underline{0.988} & \textbf{0.990} & \underline{0.913} & \textbf{0.968} & 0.970 & \underline{0.850} & \underline{0.946} & 0.880 & 23.547 & \underline{5.954}  & \textbf{0.334} & \underline{1.360}\\
\addon{\textbf{\methodname\ (ours)}} & \underline{0.988} & \textbf{0.990} & \underline{0.913} & \underline{0.979} & 0.970 & \underline{0.850} & 0.948 & \textbf{0.881} & \textbf{23.571} & \textbf{5.958}& \underline{0.334} & \textbf{1.386}\\
DiffusionOPD & \textbf{1.000} & \textbf{1.000} & \textbf{0.913} & 0.883 & 0.970 & 0.810 & 0.929 & 0.884 & 23.467 & 5.867 & 0.328 & 1.342\\
\addon{\textbf{\methodname\ (ours)}} & \textbf{1.000} & \textbf{1.000} & \textbf{0.938} & 0.957 & 0.970 & \textbf{0.860} & \textbf{0.954} & \underline{0.881} & \underline{23.558} & 5.918 & 0.333 & \underline{1.378}\\
\bottomrule
\end{tabularx}
\end{table}

\begin{figure}[!ht]
\centering
\includegraphics[width=\linewidth]{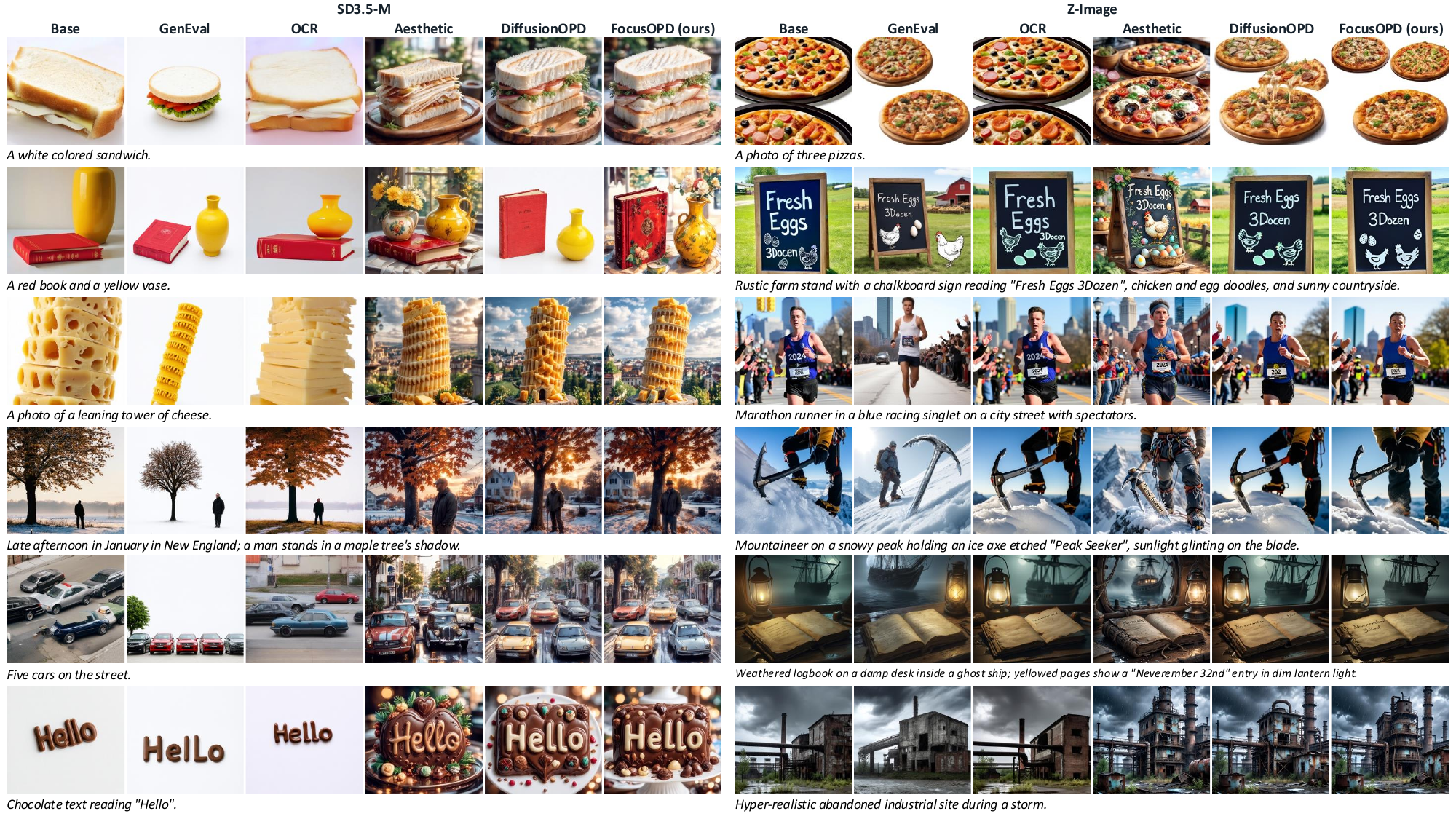}
\caption{Qualitative comparison of generation results.
Our method produces images with better compositional alignment, text rendering quality, and visual aesthetics compared to baseline methods.}
\label{fig:viz_examples}
\end{figure}

\textbf{Main-result analysis.}
We compare output-only, uniform KD, and our selective method on a fixed replay
bank, pairing depth/denoising mismatch profiles with generation scores.
This analysis tests whether better alignment in the supervised regions accompanies
better generation quality, and whether the method with the lowest global feature
error necessarily has the best task scores. Results confirm that selective
supervision in the diagnosed regions (deep layers, early denoising) improves
generation more effectively than uniform supervision across all positions.

\textbf{Cross-baseline and cross-backbone consistency.}
Our method consistently improves different OPD baselines, as shown in
Table~\ref{tab:main}. DiffusionOPD, DanceOPD, and STEP-OPD each use different
output objectives and query strategies, yet all benefit from the same Last-12
Front-40\% feature supervision on both SD3.5-M and Z-Image backbones.
This confirms that the proposed feature objective is complementary to various
output-distillation protocols.

The same table also shows that the diagnostic-guided selection strategy
generalizes from SD3.5-M to Z-Image. While both backbones use the same Last-$q$
principle (deep-layer, early-phase supervision), the specific $q$ and $r$
values are calibrated independently for each architecture through
backbone-specific diagnostics and ablations.

\subsection{Ablation Studies}
\label{sec:ablations}

All ablation studies use SD3.5-M with DiffusionOPD as the base.
We average feature loss over tokens, layers, and active steps.
Cosine distance and coefficient $\lambda$ are fixed unless varied explicitly.
Results are reported on the development prompt set with fixed generation seeds.

\begin{table}[!ht]
\centering
\caption{Ablation of feature supervision components on SD3.5-M with DiffusionOPD.
Training time per update is measured over steps 11--1000; teacher feature cache is the CPU-stored tensor payload per rank.
All feature variants use cosine distance with $\lambda=0.3$; all methods train the full 24-block transformer.}
\label{tab:component-ablation}
\scriptsize
\setlength{\tabcolsep}{3pt}
\renewcommand{\arraystretch}{1.08}
\begin{tabular}{@{}lcccccccccc@{}}
\toprule
\textbf{Configuration} & \textbf{Depth} & \textbf{Time} & \textbf{GenEval} & \textbf{OCR} & \textbf{Pick} & \textbf{Aes}& \textbf{HPS} & \textbf{ImgR} & \textbf{Time/up (s)} & \textbf{Cache (MiB)} \\
\midrule
Output KD only & -- & -- & 0.951 & \textbf{0.961} & \textbf{23.654} & 6.146 & 0.339 & \underline{1.502} & 16.54 & 0 \\
w/o depth selection & All-24 & Front 40\% & \textbf{0.965} & 0.959 & 23.565 & \underline{6.208} & \underline{0.341} & \textbf{1.505} & 19.88 & 864 \\
w/o temporal selection & Last-12 & Full & 0.951 & \underline{0.960} & 23.556 & 6.183 & \underline{0.341} & 1.500 & 19.88 & 2160 \\
All blocks, full time & All-24 & Full & 0.953 & 0.958 & \underline{23.574} & 6.198 & \textbf{0.342} & 1.501 & 21.77 & 2160 \\
\addlinespace[1pt]
\textbf{\methodname} & \textbf{Last-12} & \textbf{Front 40\%} & \underline{0.960} & 0.959 & 23.545 & \textbf{6.214} & 0.340 & 1.497 & \textbf{18.99} & \textbf{432} \\
\bottomrule
\end{tabular}
\end{table}

\textbf{Feature-selection ablation.}
Table~\ref{tab:component-ablation} examines whether depth and temporal selection are necessary for effective feature supervision.
The output-only baseline establishes that multi-teacher OPD succeeds without auxiliary features (GenEval 0.951, LAION-Aesthetic 6.146).
Adding feature supervision uniformly across all blocks and all timesteps (All-24/Full) brings modest gains in some metrics (GenEval 0.953, HPSv2.1 0.342) but requires the longest training time (21.77s per update) and largest teacher cache (2160 MiB per rank).
Comparing the two selection dimensions shows that early temporal coverage is important: supervising the full trajectory gives no GenEval improvement over output-only, whereas Front-40\% improves generation quality. For depth, All-24/Front-40\% reaches the highest GenEval (0.965) but costs more memory and time than our Last-12/Front-40\% setting, which achieves GenEval 0.960 and LAION-Aesthetic 6.214 with a smaller cache and faster updates. The detailed depth and temporal trends are analyzed below.

\begin{table}[!ht]
\centering
\begin{minipage}[t]{0.49\textwidth}
\centering
\caption{Depth window ablation (fixed Front-40\% temporal coverage). Block indices are zero-based.}
\label{tab:depth-ablation}
\scriptsize
\setlength{\tabcolsep}{2.5pt}
\renewcommand{\arraystretch}{1.05}
\resizebox{\linewidth}{!}{\begin{tabular}{@{}lcccccc@{}}
\toprule
\textbf{Depth window} & \textbf{GenEval} & \textbf{OCR} & \textbf{Pick} & \textbf{Aes}& \textbf{HPS} & \textbf{ImgR} \\
\midrule
Output KD only & 0.951 & \underline{0.961} & \textbf{23.654} & 6.146 & 0.339 & 1.502 \\
\midrule
First-6 (0--5) & 0.948 & 0.958 & \underline{23.652} & 6.207 & \textbf{0.342} & \textbf{1.513} \\
Mid-6 (8--13) & 0.958 & 0.959 & 23.616 & 6.144 & 0.340 & \underline{1.507} \\
Last-6 (18--23) & \underline{0.962} & \textbf{0.962} & 23.522 & 6.199 & 0.340 & 1.488 \\
First-12 (0--11) & 0.950 & 0.957 & 23.629 & 6.149 & \underline{0.341} & 1.506 \\
Mid-12 (6--17) & 0.953 & 0.957 & 23.607 & 6.184 & \underline{0.341} & 1.506 \\
\textbf{Last-12 (12--23)} & 0.960 & 0.959 & 23.545 & \underline{6.214} & 0.340 & 1.497 \\
Last-18 (6--23) & \underline{0.962} & 0.959 & 23.562 & 6.200 & \underline{0.341} & 1.501 \\
All blocks (0--23) & \textbf{0.965} & 0.959 & 23.565 & 6.208 & \underline{0.341} & 1.505 \\
First-6+Last-6 & \underline{0.962} & 0.959 & 23.554 & \textbf{6.225} & \underline{0.341} & 1.498 \\
\bottomrule
\end{tabular}}
\end{minipage}\hfill
\begin{minipage}[t]{0.49\textwidth}
\centering
\caption{Temporal window ablation (fixed Last-12 depth coverage). Step indices are zero-based.}
\label{tab:temporal-ablation}
\scriptsize
\setlength{\tabcolsep}{2.5pt}
\renewcommand{\arraystretch}{1.05}
\resizebox{\linewidth}{!}{\begin{tabular}{@{}lcccccc@{}}
\toprule
\textbf{Temporal window} & \textbf{GenEval} & \textbf{OCR} & \textbf{Pick} & \textbf{Aes}& \textbf{HPS} & \textbf{ImgR} \\
\midrule
Output KD only & 0.951 & \underline{0.961} & \textbf{23.654} & 6.146 & 0.339 & \textbf{1.502} \\
\midrule
Front 20\% (0--1) & 0.959 & 0.959 & 23.543 & 6.174 & 0.339 & 1.481 \\
Front 30\% (0--2) & 0.957 & 0.960 & 23.543 & \underline{6.200} & \underline{0.340} & 1.490 \\
\textbf{Front 40\% (0--3)} & \underline{0.960} & 0.959 & 23.545 & \textbf{6.214} & \underline{0.340} & 1.497 \\
Front 50\% (0--4) & \underline{0.960} & 0.960 & 23.549 & \textbf{6.214} & \textbf{0.341} & \underline{1.501} \\
Front 60\% (0--5) & 0.956 & 0.960 & 23.554 & \underline{6.200} & \textbf{0.341} & 1.500 \\
Full trajectory (0--9) & 0.951 & 0.960 & 23.556 & 6.183 & \textbf{0.341} & 1.500 \\
Middle 40\% (3--6) & 0.953 & \textbf{0.963} & \underline{23.590} & 6.174 & 0.339 & 1.481 \\
Last 40\% (6--9) & 0.954 & 0.960 & 23.579 & 6.092 & 0.335 & 1.484 \\
Front 20\%+Last 20\% & \textbf{0.963} & \underline{0.961} & 23.564 & 6.198 & \textbf{0.341} & 1.495 \\
\bottomrule
\end{tabular}}
\end{minipage}
\end{table}

\textbf{Depth selection.}
Table~\ref{tab:depth-ablation} examines which layers benefit most from feature supervision, holding temporal coverage fixed at Front-40\%.
Shallow supervision (First-6, blocks 0--5) yields minimal improvement over output-only (GenEval 0.948 vs.\ 0.951), suggesting that early-layer features are already well-aligned through output distillation.
Middle layers (Mid-6, Mid-12) show moderate gains but remain below deep-layer configurations.

Deep-layer supervision consistently outperforms shallow supervision: Last-6 (blocks 18--23) reaches GenEval 0.962 and OCR 0.962, the highest OCR among all depth windows.
Expanding to Last-12 (blocks 12--23) achieves a LAION-Aesthetic score of 6.214 while maintaining strong compositional performance (GenEval 0.960).
Further expansion to Last-18 or All-24 increases GenEval slightly (0.962, 0.965) but at diminishing returns: All-24 requires double the feature computation of Last-12 with only +0.5\% GenEval gain.
Non-contiguous selection (First-6+Last-6) does not outperform Last-12, confirming that a contiguous deep window suffices.
These results support the hypothesis that mismatch concentrated in deeper blocks benefits most from explicit alignment, as these representations are more task-specific and further from the shared backbone prior.

\textbf{Temporal selection.}
Table~\ref{tab:temporal-ablation} varies the denoising-step coverage with Last-12 depth fixed.
Supervising the full trajectory (steps 0--9) provides no gain over output-only (GenEval 0.951 for both), despite supervising 2.5$\times$ more layer-step pairs than Front-40\%.
Middle 40\% (steps 3--6) and Last 40\% (steps 6--9) similarly underperform, with Last-40\% yielding the lowest LAION-Aesthetic (6.092) and HPSv2.1 (0.335) among all temporal configurations.

In contrast, early-phase supervision drives consistent improvements.
Front-20\% (steps 0--1) already lifts GenEval from 0.951 to 0.959.
Extending to Front-40\% (steps 0--3) reaches GenEval 0.960 and LAION-Aesthetic 6.214, the best aesthetic score across all temporal windows.
Front-50\% matches this GenEval and LAION-Aesthetic, with marginal HPSv2.1 gain (+0.001), but extending further to Front-60\% or Full trajectory degrades performance.
The sharp drop from Front-50\% (GenEval 0.960) to Full (0.951) indicates that late denoising steps add no value to feature supervision and may even dilute the training signal from high-noise steps.
This occurs because noise levels are low and model priors converge at late steps.
This empirically confirms the prediction that feature mismatch matters most when noise modulates sensitivity to prior differences, precisely the regime our theory identifies.
\textbf{Feature-loss formulation.}
Table~\ref{tab:loss-type} compares cosine distance with teacher-energy-normalized
MSE under the selected Last-12 and Front-40\% windows. Cosine distance gives
the strongest GenEval and LAION-Aesthetic scores, supporting its use as the default feature
loss.

\begin{table}[!ht]
\centering
\caption{Feature loss formulations on SD3.5-M with Last-12 and front-40\% coverage.
Both feature variants use $\lambda=0.3$.
Normalized MSE divides each sample/layer squared error by teacher mean squared energy,
then averages over selected layers; the denominator floor is $10^{-8}$.
This compares formulations at a fixed coefficient, not matched gradient magnitudes.}
\label{tab:loss-type}
\small
\begin{tabular}{@{}lcccccc@{}}
\toprule
Feature loss type & GenEval & OCR & PickScore & Aes  & HPSv2.1 & ImgRwd \\
\midrule
Output KD only (no feature) & 0.951 & \underline{0.961} & \textbf{23.654} & 6.146 & \underline{0.339} & \textbf{1.502} \\
Teacher-energy-normalized MSE & \underline{0.955} & \textbf{0.963} & 23.509 & \underline{6.210} & \textbf{0.340} & 1.491 \\
Cosine distance & \textbf{0.960} & 0.959 & \underline{23.545} & \textbf{6.214} & \textbf{0.340} & \underline{1.497} \\
\bottomrule
\end{tabular}
\end{table}

\section{Related Work}
\label{sec:related_work}
We provide a detailed discussion of related work in Appendix~\ref{app:full_related_work}. On-policy distillation (OPD) trains students on their own trajectories, reducing the distribution mismatch between training and inference \citep{agarwal2024onpolicy}. Recent studies \citep{diffusionopd,flowopd,danceopd,dopsd} extend OPD to multi-teacher capability integration in diffusion models, but primarily rely on output-level supervision. Feature distillation \cite{romero2015fitnets,lee2024koala,yu2025repa,stepopd} adds intermediate representation supervision, while selective knowledge distillation \citep{liang2023ted,liu2024dfd} adapts the supervised content or location according to teacher relevance. However, how to select feature supervision across network depth and denoising stages in multi-teacher diffusion OPD remains underexplored. Building on output-level OPD, our method supervises conditional image-token representations in selected deep blocks during early denoising to study their depth- and stage-dependent benefits.

\section{Conclusion}
In this work, we investigate where and when intermediate supervision is most effective for multi-teacher on-policy distillation in diffusion models. We identify a persistent depth–temporal structure in teacher–student mismatch and provide theoretical and empirical evidence linking it to selective supervision. Building on these findings, we introduce FocusOPD, which complements output-level distillation with directional feature alignment in deep layers during early denoising. Experiments across two diffusion backbones and multiple OPD baselines show consistent gains, demonstrating the effectiveness of structure-guided feature supervision for multi-teacher capability integration.

\section*{AI Use Statement}
In this work, we used generative AI tools to assist with translation, polish and clarify the writing, and support the retrieval and discovery of related work. We also used them to support environment construction, testing, and monitoring in our experiments. Specifically, we used large language models (LLMs), including GPT-6 and Claude Opus 5, to improve the clarity and readability of the manuscript. We have reviewed all AI-assisted work and take full responsibility for the final content of this work, including any text, claims, or artifacts produced with the assistance of generative AI.
\bibliography{references}

@article{diffusionopd,
  author={Li, Quanhao and others},
  title={{DiffusionOPD}: A Unified Perspective of On-Policy Distillation in Diffusion Models},
  journal={arXiv preprint arXiv:2605.15055}, year={2026}}

@article{sd3,
  author={Esser, Patrick and others},
  title={Scaling Rectified Flow Transformers for High-Resolution Image Synthesis},
  journal={arXiv preprint arXiv:2403.03206}, year={2024}}

@article{zimage,
  author={{Z-Image Team} and others},
  title={{Z-Image}: An Efficient Image Generation Foundation Model with Single-Stream Diffusion Transformer},
  journal={arXiv preprint arXiv:2511.22699}, year={2025}}

@inproceedings{hu2022lora,
  author={Hu, Edward J. and Shen, Yelong and Wallis, Phillip and Allen-Zhu, Zeyuan and Li, Yuanzhi and Wang, Shean and Wang, Lu and Chen, Weizhu},
  title={LoRA: Low-Rank Adaptation of Large Language Models},
  booktitle={International Conference on Learning Representations}, year={2022}}

@article{geneval,
  author={Ghosh, Dhruba and Hajishirzi, Hannaneh and Schmidt, Ludwig},
  title={{GenEval}: An Object-Focused Framework for Evaluating Text-to-Image Alignment},
  journal={Advances in Neural Information Processing Systems}, year={2023}}

@article{imagen,
  author={Saharia, Chitwan and others},
  title={Photorealistic Text-to-Image Diffusion Models with Deep Language Understanding},
  journal={Advances in Neural Information Processing Systems}, year={2022}}

@article{pickscore,
  author={Kirstain, Yuval and others},
  title={Pick-a-Pic: An Open Dataset of User Preferences for Text-to-Image Generation},
  journal={Advances in Neural Information Processing Systems}, year={2023}}

@misc{hps,
  author={Wu, Xiaoshi and others}, title={Human Preference Score v2}, year={2023},
  howpublished={\url{https://github.com/tgxs002/HPSv2}}, note={Evaluation checkpoint: v2.1}}

@misc{aesthetic,
  author={Schuhmann, Christoph}, title={{LAION-Aesthetics}}, year={2022},
  howpublished={\url{https://laion.ai/blog/laion-aesthetics/}}}

@article{imagereward,
  author={Xu, Jiazheng and others},
  title={{ImageReward}: Learning and Evaluating Human Preferences for Text-to-Image Generation},
  journal={Advances in Neural Information Processing Systems}, year={2023}}

@article{danceopd,
  author={Zhou, Wei and Zhu, Xiongwei and Xu, Zelin and Dong, Bo and Gong, Lixue and Liang, Yongyuan and Chu, Meng and Qu, Leigang and Kong, Lingdong and Liu, Wei and Chua, Tat-Seng},
  title={{DanceOPD}: On-Policy Generative Field Distillation},
  journal={arXiv preprint arXiv:2606.27377}, year={2026}}

@article{stepopd,
  author={Wei, Qingyan and Li, Guangzhao and Tu, Xiaobing and Wang, Yinggui and Zhang, Xiantao and Ren, Jinkui and Liu, Xiaohong and Zhang, Linfeng},
  title={{STEP-OPD}: Rethinking Output Targets and Internal Dynamics in On-Policy Distillation for Diffusion Models},
  journal={arXiv preprint arXiv:2608.04887}, year={2026}}

@article{flowopd,
  author={Fang, Zhen and Huang, Wenxuan and Zeng, Yu and Zhao, Yiming and Chen, Shuang and Feng, Kaituo and Lin, Yunlong and Chen, Lin and Chen, Zehui and Cao, Shaosheng and Zhao, Feng},
  title={{Flow-OPD}: On-Policy Distillation for Flow Matching Models},
  journal={arXiv preprint arXiv:2605.08063}, year={2026}}

@article{dopsd,
  author={Jiang, Dengyang and Jin, Xin and Liu, Dongyang and Wang, Zanyi and Zheng, Mingzhe and Du, Ruoyi and Yang, Xiangpeng and Wu, Qilong and Li, Zhen and Gao, Peng and Yang, Harry and Hoi, Steven},
  title={{D-OPSD}: On-Policy Self-Distillation for Continuously Tuning Step-Distilled Diffusion Models},
  journal={arXiv preprint arXiv:2605.05204}, year={2026}}

@inproceedings{agarwal2024onpolicy,
  title = {On-Policy Distillation of Language Models: Learning from Self-Generated Mistakes},
  author = {Agarwal, Rishabh and Vieillard, Nino and Zhou, Yongchao and Stanczyk, Piotr and Ramos Garea, Sabela and Geist, Matthieu and Bachem, Olivier},
  booktitle = {The Twelfth International Conference on Learning Representations},
  year = {2024},
  url = {https://proceedings.iclr.cc/paper_files/paper/2024/hash/5be69a584901a26c521c2b51e40a4c20-Abstract-Conference.html}
}

@article{lu2025onpolicydistillation,
  author = {Lu, Kevin and {Thinking Machines Lab}},
  title = {On-Policy Distillation},
  journal = {Thinking Machines Lab: Connectionism},
  year = {2025},
  url = {https://thinkingmachines.ai/blog/on-policy-distillation/},
  doi = {10.64434/tml.20251026}
}

@article{ma2026mopd,
  title = {{MOPD}: Multi-Teacher On-Policy Distillation for Capability Integration in {LLM} Post-Training},
  author = {Ma, Wenhan and Wei, Jianyu and Zhao, Liang and Zhang, Hailin and Xiao, Bangjun and Li, Lei and Yang, Qibin and Gao, Bofei and Wang, Yudong and Li, Rang and Dong, Jinhao and Sui, Zhifang and Luo, Fuli},
  journal = {arXiv preprint arXiv:2606.30406},
  year = {2026},
  url = {https://arxiv.org/abs/2606.30406}
}

@inproceedings{romero2015fitnets,
  title = {{FitNets}: Hints for Thin Deep Nets},
  author = {Romero, Adriana and Ballas, Nicolas and Kahou, Samira Ebrahimi and Chassang, Antoine and Gatta, Carlo and Bengio, Yoshua},
  booktitle = {International Conference on Learning Representations},
  year = {2015},
  url = {https://arxiv.org/abs/1412.6550}
}

@inproceedings{lee2024koala,
  title = {{KOALA}: Empirical Lessons Toward Memory-Efficient and Fast Diffusion Models for Text-to-Image Synthesis},
  author = {Lee, Youngwan and Park, Kwanyong and Cho, Yoorhim and Lee, Yong-Ju and Hwang, Sung Ju},
  booktitle = {Advances in Neural Information Processing Systems},
  volume = {37},
  year = {2024},
  url = {https://proceedings.neurips.cc/paper_files/paper/2024/hash/5c4e0e38691e2aa08bba4cefc4c6e852-Abstract-Conference.html}
}

@inproceedings{yu2025repa,
  title = {Representation Alignment for Generation: Training Diffusion Transformers Is Easier Than You Think},
  author = {Yu, Sihyun and Kwak, Sangkyung and Jang, Huiwon and Jeong, Jongheon and Huang, Jonathan and Shin, Jinwoo and Xie, Saining},
  booktitle = {The Thirteenth International Conference on Learning Representations},
  year = {2025},
  url = {https://proceedings.iclr.cc/paper_files/paper/2025/hash/d9e42b4d7163931f3689d6d6fbaa11d0-Abstract-Conference.html}
}

@inproceedings{liang2023ted,
  title = {Less is More: Task-aware Layer-wise Distillation for Language Model Compression},
  author = {Liang, Chen and Zuo, Simiao and Zhang, Qingru and He, Pengcheng and Chen, Weizhu and Zhao, Tuo},
  booktitle = {Proceedings of the 40th International Conference on Machine Learning},
  series = {Proceedings of Machine Learning Research},
  volume = {202},
  pages = {20852--20867},
  year = {2023},
  publisher = {PMLR},
  url = {https://proceedings.mlr.press/v202/liang23j.html}
}

@inproceedings{liu2024dfd,
  title = {{DFD}: Distilling the Feature Disparity Differently for Detectors},
  author = {Liu, Kang and Zhang, Yingyi and Zhang, Jingyun and Li, Jinmin and Wang, Jun and Wang, Shaoming and Yuan, Chun and Guo, Rizen},
  booktitle = {Proceedings of the 41st International Conference on Machine Learning},
  series = {Proceedings of Machine Learning Research},
  volume = {235},
  pages = {32421--32430},
  year = {2024},
  publisher = {PMLR},
  url = {https://proceedings.mlr.press/v235/liu24cd.html}
}
\bibliographystyle{iclr2027_conference}
\clearpage
\appendix

\section{Proofs of the Mismatch and Supervision Results}
\label{app:mismatch_theory}

\subsection{Depth-wise Discrepancy Transport}
\label{app:discrepancy_propagation}

\begin{proof}[Proof of Lemma~\ref{lem:discrepancy_propagation}]
Fix $s$ and suppress its index. Adding and subtracting
$F_\ell^S(z_\ell^T)$ and applying the fundamental theorem of calculus gives
\begin{equation}
    d_{\ell+1}=b_\ell+A_\ell d_\ell,
    \qquad
    A_\ell=\int_0^1 JF_\ell^S(z_\ell^T+u d_\ell)\,du,
    \label{eq:discrepancy_transport}
\end{equation}
where $J$ denotes the input Jacobian. Recursively substituting this
identity gives Equation~\eqref{eq:discrepancy_expansion}, including the
initial embedded-state difference $d_0$.
\end{proof}

For the diagnostic in Section~\ref{sec:validation}, let
$p_\ell=A_\ell d_\ell=F_\ell^S(z_\ell^S)-F_\ell^S(z_\ell^T)$
and let $\Pi$ select image-token coordinates at the block output. Then
\begin{equation}
    \|\Pi d_{\ell+1}\|_2^2
    =\|\Pi b_\ell\|_2^2+\|\Pi p_\ell\|_2^2
     +2\langle\Pi b_\ell,\Pi p_\ell\rangle.
    \label{eq:discrepancy_cross_term}
\end{equation}
The cross term permits reinforcement or cancellation, so the transport
identity alone does not imply monotonic growth with depth. Shared noisy
latents also need not give $d_0=0$ when input embeddings differ.

\subsection{Noise-dependent Prior Sensitivity}
\label{app:prior_sensitivity}

\begin{proof}[Proof of Lemma~\ref{lem:prior_sensitivity}]
Gaussian conditioning yields
\begin{equation}
    \widehat y_M
    =\frac{a_s\tau^2}{a_s^2\tau^2+\sigma_s^2}x_s
     +\frac{\sigma_s^2}{a_s^2\tau^2+\sigma_s^2}\mu_M(c).
\end{equation}
At the same $(x_s,c)$, subtracting the two estimates cancels the
observation term and gives Equation~\eqref{eq:prior_retention}.
Since $\rho_s=(1+\tau^2\mathrm{SNR}_s)^{-1}$ decreases with SNR,
weaker observations preserve more of the prior-mean difference.
\end{proof}

The shared covariance is essential to this scalar response. The lemma
models clean-latent estimates; it does not impose the same response on
hidden features or velocity predictions. In the flow parameterization
$x_t=(1-t)y+t\varepsilon$ with
$\widehat v_M=(x_t-\widehat y_M)/t$, for $t>0$,
\begin{equation}
    \widehat v_S-\widehat v_T
    =-\frac{t}{\tau^2(1-t)^2+t^2}\bigl(\mu_S(c)-\mu_T(c)\bigr).
\end{equation}
This coefficient need not grow monotonically with noise. The following
update analysis therefore uses the actual feature loss directly.

\subsection{Local Utility of Selective Supervision}
\label{app:selective_update}

\begin{proof}[Proof of Proposition~\ref{prop:selective_update}]
All gradients defining the updates are computed at the same $\theta$.
If $D_i=0$, differentiability gives
$g_i=J_{e_i}(\theta)^\top e_i(\theta)=0$, justifying $\alpha_i=0$.
For $u=\theta_{\mathrm{out}}^+$, the Lipschitz-gradient bound gives
\begin{equation}
    \begin{aligned}
        \mathcal L_{\mathrm{out}}(u-\delta g_p)
        &=\mathcal L_{\mathrm{out}}(u)
          -\delta\langle g_{\mathrm{out}}^+,g_p\rangle+r_p,\\
        |r_p|&\leq\frac{L_{\mathrm{sm}}\delta^2}{2}\|g_p\|_2^2.
    \end{aligned}
\end{equation}
Substitute $g_p=\sum_i p_i g_i$ and
$\langle g_{\mathrm{out}}^+,g_i\rangle=\alpha_iD_i$ to obtain
Equation~\eqref{eq:selective_output_gain}, with $R_p=-r_p$.
For two distributions obeying the compatibility bounds, let
$\Delta=\alpha_{\min}S_p-\alpha_{\max}S_q>0$. Then
\begin{equation}
    \mathcal G(p)-\mathcal G(q)
    \geq\delta\Delta
       -\frac{L_{\mathrm{sm}}\delta^2}{2}
        \bigl(\|g_p\|_2^2+\|g_q\|_2^2\bigr).
    \label{eq:selective_comparison_bound}
\end{equation}
The right-hand side is positive for
$0<\delta<2\Delta/
[L_{\mathrm{sm}}(\|g_p\|_2^2+\|g_q\|_2^2)]$ when the denominator
is positive, and for any positive $\delta$ within the smoothness
neighborhood when it is zero. Here $\eta$ is fixed while the common
auxiliary weight controls $\delta$.
\end{proof}

\paragraph{Interpretation of the assumptions.}
Replay states, teacher targets, and the candidate distributions are
fixed only for this comparison. The argument uses ordinary Euclidean
gradient steps and requires neither a teacher-matching parameter
reference nor a quadratic loss or stable training trajectory.
The lemmas motivate the mismatch structure; positive and comparable
$\alpha_i$ are an additional local compatibility condition.
Equal $\sum_i p_i=\sum_i q_i=1$ fixes coefficient mass, not gradient norm.

The gradient $g_{\mathrm{out}}^+$ is evaluated at the output-only
comparison point, while every feature gradient remains evaluated at
$\theta$; no sequential recomputation is required by the joint update.
Replacing $g_{\mathrm{out}}^+$ by $g$ would add an error bounded by
$L_{\mathrm{sm}}\eta\delta\|g\|_2\|g_p\|_2$, provided the same
smoothness bound also holds between $\theta$ and
$\theta_{\mathrm{out}}^+$. This is first order in $\delta$ at fixed
$\eta$. The result supplies a local allocation principle; persistence
of the selected region remains an empirical observation.

\subsection{Direct Connection to Cosine Supervision}
\label{app:angular_response}

Let $u_k^M=h_k^M/\|h_k^M\|_2$ be unit-normalized image tokens at
position $i$, and let $Q$ be the total number of tokens across the
fixed batch. Stacking their differences as
$e_i=Q^{-1/2}\operatorname{vec}(U^S-U^T)$ gives the exact identity
\begin{equation}
    \frac12D_i
    =\frac1Q\sum_{k=1}^{Q}
       \bigl(1-\langle u_k^S,u_k^T\rangle\bigr)
    =\ell_{\mathrm{cos},i}.
    \label{eq:cosine_quadratic}
\end{equation}
Proposition~\ref{prop:selective_update} thus applies to the actual
cosine-loss gradients, without a feature-response linearization.
Its mismatch variable is angular discrepancy; a raw-MSE ranking does
not by itself establish the same ordering.
The identity assumes nonzero tokens with the numerical norm floor
inactive. Where the safeguarded loss is differentiable but this identity
does not hold, the general gain expansion still applies directly to its
loss gradient through $\delta\sum_i p_i
\langle g_{\mathrm{out}}^+,\nabla\ell_i\rangle+R_p$.

\section{Full Related Work.}
\label{app:full_related_work}

\paragraph{On-policy distillation for capability integration.} On-policy distillation (OPD) \citep{agarwal2024onpolicy} transfers teacher knowledge on student-generated trajectories, addressing the distribution mismatch between training and
inference. Recent work \citep{ma2026mopd,diffusionopd,flowopd} extends this paradigm to capability integration from multiple teachers in language and diffusion models. DiffusionOPD \cite{diffusionopd} derives transition-level supervision for diffusion processes, while Flow-OPD \cite{flowopd} combines task-specialized teachers with a task-agnostic anchor for multi-task alignment. DanceOPD \cite{danceopd} formulates capability transfer as generative field distillation, using sample-wise routing and a single low-noise velocity query. D-OPSD \cite{dopsd} applies on-policy self-distillation to continual adaptation of few-step models. Building on output-level OPD, our work investigates intermediate teacher supervision for multi-task capability transfer.

\paragraph{Feature alignment in diffusion models.}
Feature distillation supplements output matching with
intermediate teacher representations \citep{romero2015fitnets}. In diffusion models, KOALA \cite{lee2024koala} studies self-attention distillation for compact text-to-image generators, while REPA \cite{yu2025repa} aligns denoising features with pretrained visual representations to improve diffusion-transformer training. STEP-OPD \citep{stepopd} introduces
representation-change alignment alongside output extrapolation, matching the direction and magnitude of inter-block updates. Our method directly aligns conditional image-token representations within selected deep blocks and an early denoising window, with feature supervision scheduled independently of output queries.

\paragraph{Selective and adaptive knowledge distillation.}
Selective distillation accounts for differences in
the relevance and transferability of teacher knowledge. TED \cite{liang2023ted} learns task-aware filters for intermediate representations in language-model compression. DFD \cite{liu2024dfd} assigns different distillation constraints to spatial regions according to teacher–student feature disparity in object detection. SAKD learns a routing policy to adapt distillation locations to individual samples and training iterations. These methods select feature content, spatial constraints, or distillation locations in discriminative tasks. We study which blocks and denoising stages benefit from additional feature supervision in diffusion OPD.
\section{Full results of Empirical Evidence for the Mismatch Mechanisms}
\label{app:evidence}
\begin{figure}
    \centering
    \includegraphics[width=0.9\linewidth]{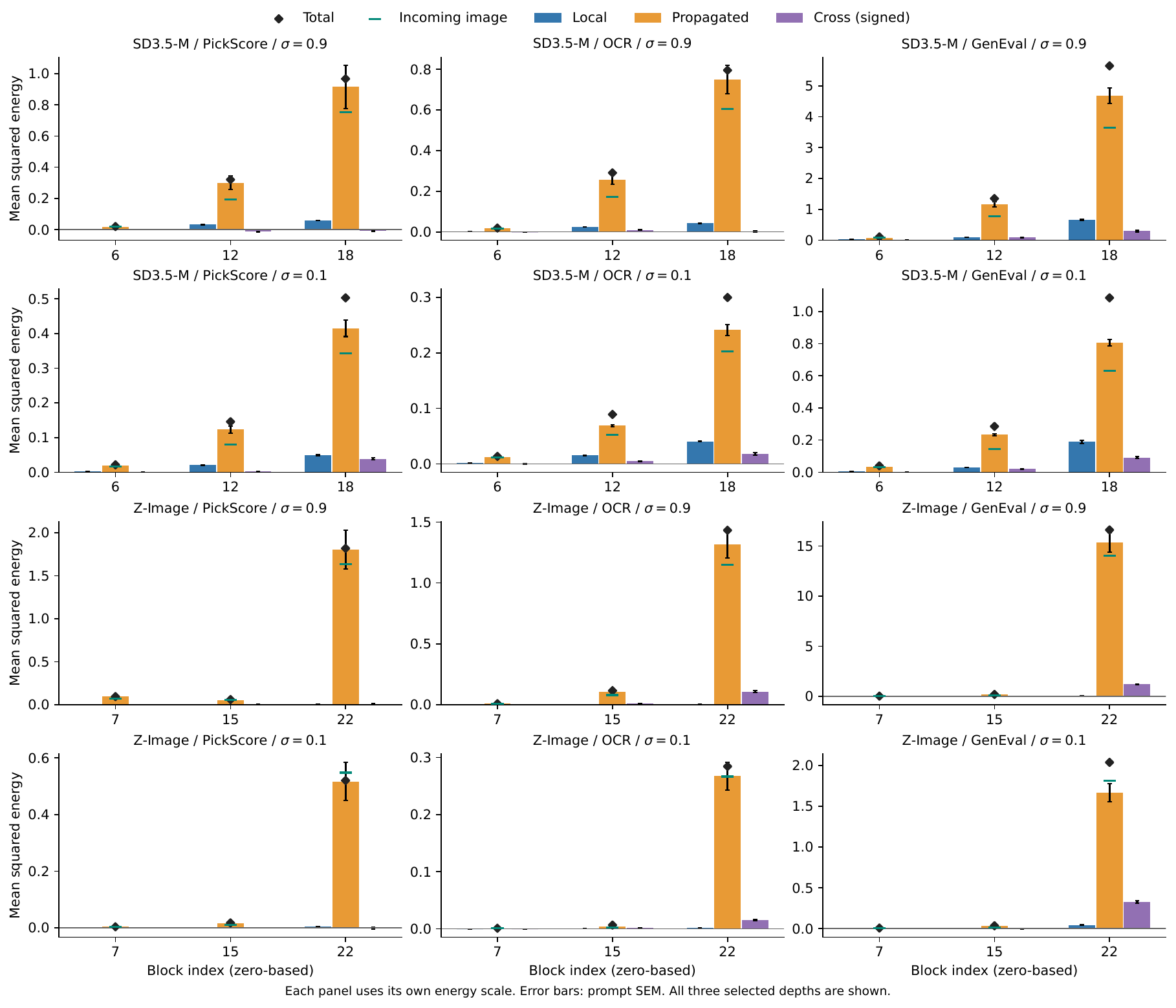}
    \caption{Deep-block decomposition across three teachers for all tested $\sigma$. Terms are
        normalized by total mean discrepancy energy. Block indices are
        18 for SD3.5-M and 22 for Z-Image.
        Error bars: 95\% prompt-bootstrap intervals.}
    \label{fig:all_block}
\end{figure}

Figure~\ref{fig:all_block} presents the full version of Figure~\ref{fig:depth-decomposition}. As shown, the cumulative effect dominates the teacher–student mismatch across depths, stages, and configurations of different networks.
\end{document}